# Carbon Footprint Reduction for Sustainable Data Centers in Real-Time

**Soumyendu Sarkar[*†], Avisek Naug[†], Ricardo Luna[†], Antonio Guillen[†], Vineet Gundecha[†], Sahand Ghorbanpour, Sajad Mousavi, Dejan Markovikj, Ashwin Ramesh Babu**

Hewelett Packard Labs @ Hewlett Packard Enterprise
820 N McCarthy Blvd, Milpitas, CA 95035, USA
soumyendu.sarkar, avisek.naug, rluna, antonio.guillen, vineet.gundecha, sahand.ghorbanpour, sajad.mousavi, dejan.markovikj, ashwin.ramesh-babu@hpe.com

## Abstract

As machine learning workloads significantly increase energy consumption, sustainable data centers with low carbon emissions are becoming a top priority for governments and corporations worldwide. This requires a paradigm shift in optimizing power consumption in cooling and IT loads, shifting flexible loads based on the availability of renewable energy in the power grid, and leveraging battery storage from the uninterrupted power supply in data centers, using collaborative agents. The complex association between these optimization strategies and their dependencies on variable external factors like weather and the power grid carbon intensity makes this a hard problem. Currently, a real-time controller to optimize all these goals simultaneously in a dynamic real-world setting is lacking. We propose a Data Center Carbon Footprint Reduction (DC-CFR) multi-agent Reinforcement Learning (MARL) framework that optimizes data centers for the multiple objectives of carbon footprint reduction, energy consumption, and energy cost. The results show that the DC-CFR MARL agents effectively resolved the complex interdependencies in optimizing cooling, load shifting, and energy storage in real-time for various locations under real-world dynamic weather and grid carbon intensity conditions. DC-CFR significantly outperformed the industry-standard ASHRAE controller with a considerable reduction in carbon emissions (14.5%), energy usage (14.4%), and energy cost (13.7%) when evaluated over one year across multiple geographical regions.

## Introduction

In recent years, sustainability and carbon footprint reduction have emerged as critical factors driving the need for innovative optimization techniques in data center (DC) operations. While energy and cost optimization have been primary concerns in smart-grid problems, the increasing sustainability commitments of companies with large DCs have made carbon footprint reduction an essential target for the industry. Achieving significant carbon footprint savings requires reducing energy consumption and replacing carbon-intensive energy sources with those with a lower carbon footprint.

Static, isolated approaches for carbon footprint reduction, such as energy optimization, load shifting to less carbon-intensive hours, and battery usage for charging during low **power grid carbon intensity (CI)** hours to supplement load demand during high CI hours, are frequently used. However, achieving significant footprint savings with analytic pipeline-based planning has proven challenging due to the complexity of these individual problems and the reliance on long forecast horizons (24h) for static approaches. Furthermore, the dependencies between these approaches and the necessity of information exchange across separate problems have prevented the development of a cohesive strategy that can simultaneously reduce the carbon footprint using all three methods in real time.

In this paper, we introduce DC Carbon Footprint Reduction (DC-CFR), a novel framework that uses Multi-Agent Deep Reinforcement Learning (DRL) to optimize DC energy consumption, flexible load shifting, and battery operation decisions simultaneously in real time. The optimization is based on short-term weather and grid CI information. Grid CI refers to the amount of $CO_2$ emissions produced per unit of electricity consumed, which varies based on the source of the electricity (e.g., fossil fuels, renewable energy) at a given time. The lower the CI, the more renewable the energy source. Our approach effectively mitigates the drawbacks of existing, isolated methods. It does so by efficiently managing the complex interdependencies and information exchange among individual optimization strategies, a process that is illustrated in Figure 2 at a system level and in Figure 3 to show the dependencies.

The proposed contributions of the framework are as follows:

- A carbon emission-aware framework for controlling data centers by redistribution of server workloads, efficient cooling, and battery storage for auxiliary energy supply.
- Real-time control for the individual approaches under the framework, while coordinating between themselves using shared reward and state variables. The collaborative performance indicators help the agents self-adjust their operations.
- Implementation of the framework as a multi-agent reinforcement learning problem using industry-standard simulators for Load Shifting and Battery Supply from Meta (2) and Energy Plus software from NREL (6) with a Sinergym Wrapper (9).

[*]Corresponding author.
[†]These authors contributed equally.

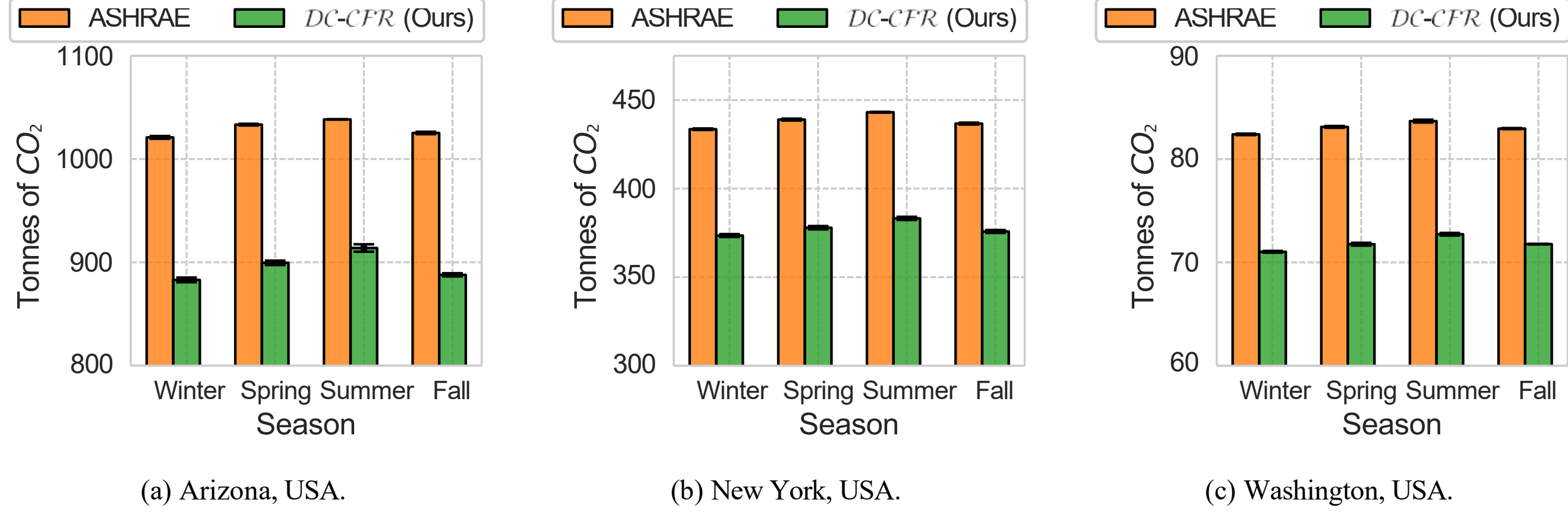

(a) Arizona, USA. (b) New York, USA. (c) Washington, USA.

Figure 1: $CO_2$ generation (Tonnes) for data center (DC) control approaches in a 1.2 MWh DC in different locations. ASHRAE is an industry-standard controller for HVAC in DCs.

- Extensive evaluation of the approach across multiple geographical locations with different weather patterns and benchmark it against the current industry standard ASHRAE rule-based-controller (RBC).

Our approach can significantly decrease carbon emissions in data centers in the different tested locations (refer to Figure 1 for detailed results). Over a span of one year, and across different DC configurations and geographical locations, the DC-CFR framework demonstrated an average carbon footprint reduction of 14.46%.

In addition to reducing carbon emissions, our evaluations revealed the DC-CFR system's capacity to curtail energy consumption. On average, our tests showed a reduction in energy usage by 14.35%. Furthermore, the DC-CFR system also efficiently managed energy costs, reducing the average energy expenditure by 13.69%. These results underscore the potential of DC-CFR as a comprehensive and effective solution for achieving sustainability goals in DC operations.

## Related Work

### Energy Savings

Deep Reinforcement Learning (DRL) shows promise in dynamic thermal management in DCs, specifically for reducing energy consumption via Heating, Ventilation, and Air Conditioning (HVAC) system control (43; 42; 13; 3; 39; 15; 16). However, the real-world deployment of DRL-based systems is complicated by their sensitivity to hyperparameters, reward functions, and work scenarios (43; 13; 3; 39; 29; 27; 30; 31; 28; 36; 21; 22; 25; 33; 24; 23; 38). Moreover, ensuring safety and satisfying operational constraints, especially for HVAC system control, is another challenge (43; 39).

Despite the challenges, DRL has shown potential for energy savings in DCs. DRL-based strategies have achieved up to 12% savings compared to default controllers (43), and 8.84% compared to reference controllers. Additional savings of up to 5.5% have been noted in tropical climates (13), while in simulated environments, a reduction of at least 10% in energy consumption has been observed (3).

### Load Shifting and Battery Optimization

With DCs accounting for a significant portion of global energy consumption, Carbon-Aware Workload Scheduling (CAS) has emerged as a potential solution (2; 17). CAS uses delay-tolerant workloads to decrease carbon emissions by rescheduling them to times of lower CI. For instance, the Carbon Explorer framework (2) reduces the overall DC footprint by $\sim$ 4% on historical data by shifting the flexible part of the DC load to the lowest carbon-intensive hours.

DRL has been applied to optimize workload scheduling in DCs, improving energy efficiency (19; 18; 20; 40). One approach, GreenDRL, uses DRL for CAS, showing a reduction in the energy obtained from the main grid and an increase in the use of green energy (41). However, GreenDRL primarily considers scenarios with on-site renewable energy resources.

Battery operation optimization is another area of focus, with strategies divided into static schedules based on day-ahead information and real-time control for when longer-term forecasts are unreliable (2; 17).

Real-time battery optimization strategies using DRL have been developed, but most overlook the degradation of battery charging and discharging rates across their instantaneous states of charge (44; 8; 1; 4). For instance, a DRL agent for optimal battery operation assuming a battery degradation model has been developed, reducing net energy costs compared to a baseline battery operation algorithm (4).

### Our Approach

While current carbon-reduction approaches show promise, they lack real-time operation capabilities and do not effectively combine multiple control strategies due to the complex interdependencies and balancing objectives. Our approach partially decouples the problem into sub-problems, each solved with an individual Markov decision process (MDP) formulation, the mathematical framework for RL, while the combined rewards and overlapping state variables in a collaborative multi-agent setting solve the dependencies in real-time. This design results in a comprehensive, real-time carbon footprint optimizer for sustainable societies, advancing beyond existing strategies to offer a more adaptable and robust solution.

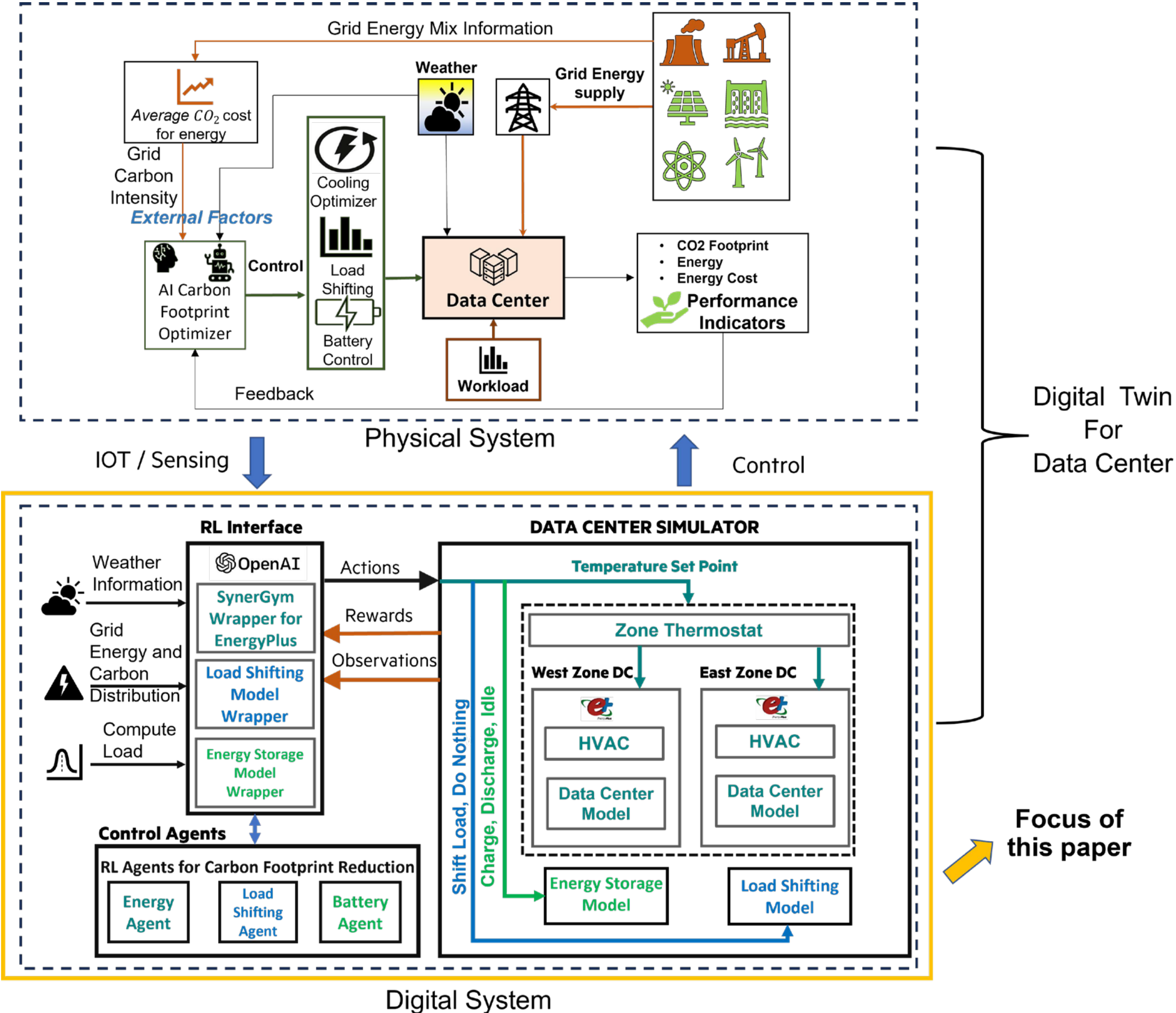

Figure 2: Overview of the physical and digital systems. For this work, we have used the simulation with EnergyPlus data center simulation from NREL, extended the RL interface with IBM's SinerGym, and used the battery model from Facebook.

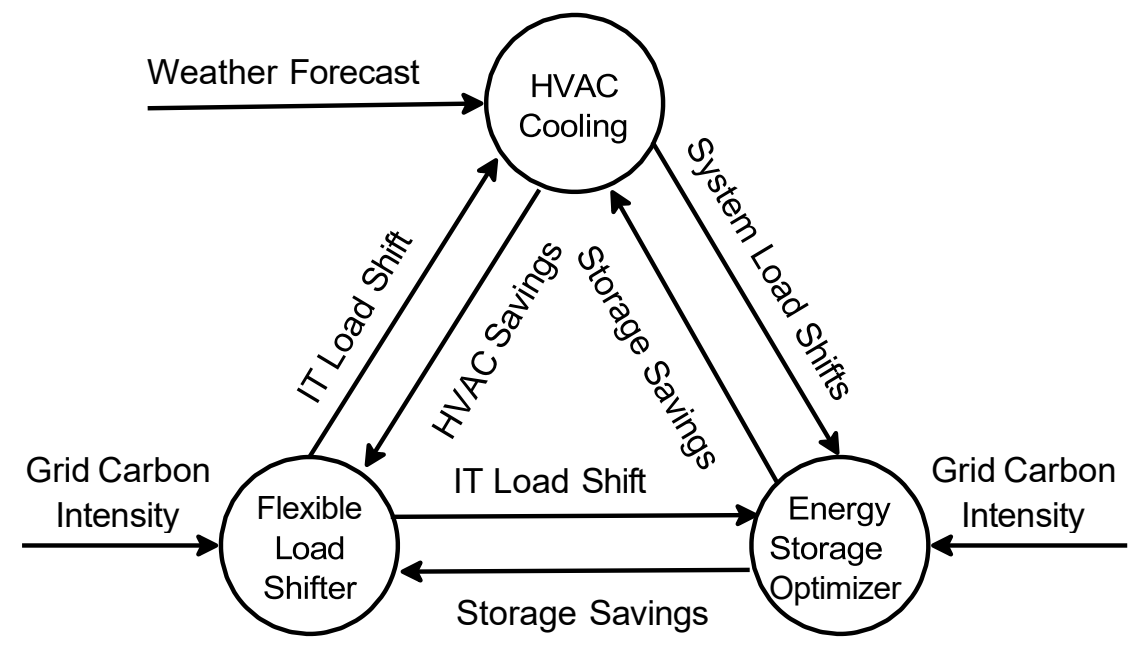

Figure 3: Internal and External Dependencies for the agents.

## Problem Definition Using Markov Decision Processes

In this section, we present the problem formulation for reducing energy and carbon footprint in data centers. We do this by considering three MDPs that take into account DC workload shifting, energy reduction through cooling setpoint optimization, and auxiliary energy supply using energy storage systems. As mentioned earlier, these three problems have been tackled individually in the literature using static approaches (2; 17) with day-ahead forecast information. We reformulate the problems so that they can be solved in real-time.

The three MDPs are described in Table 1 with their interdependencies summarized in Figure 3. Solving $\text{MDP}_{LS}$ reduces the carbon footprint and energy consumption by shifting data center workloads to low-CI and low ambient temperature hours; $\text{MDP}_{E}$ reduces data center energy use by optimizing the HVAC cooling setpoint; and $\text{MDP}_{BAT}$ reduces the carbon footprint by charging and discharging the battery based on the grid CI.

As can be seen from the MDP description, the system involves a chain of dependencies starting with the *Flexible*

*Load Shifter*, then proceeding through *HVAC Cooling*, and finally, Energy Storage Optimizer. The Load Shifter determines load shifts based on workload, grid CI, DC power usage, and thermal state. HVAC uses the resulting server workload and other factors like weather and battery charge to optimize energy with a surrogate model based on Energy Plus (9). This model estimates future energy use. The Energy Storage Optimizer uses this data and grid CI to decide battery actions. CO2Footprint rewards are calculated using CIt and TotalEnergyConsumptiont at each time step.

The interlinked dependencies offer potential energy and carbon savings but pose challenges for DRL agent training convergence. Challenges include changing state distributions with policy updates, different time constants for MDP changes, and state interdependencies. A collaborative reward framework is essential for agents to appropriately incorporate shared state variables from other MDPs into their rewards.

## Multi-Agent Reinforcement Learning

Solving the challenges of sustainable data center operation requires addressing multiple interdependent sub-problems. Multi-Agent Reinforcement Learning (MARL) is a suitable approach due to the agents' cooperative nature. This allows individual agents to pursue individual objectives while considering other agents' actions through a reward mechanism, aiding collaboration. The study explores two MARL methods. We implemented Multi-Agent Deep Deterministic Policy Gradient (MADDPG) (12), which involves decentralized agents learning from a centralized critic that incorporates all agents' behaviors. We also adapted the Independent Proximal Policy Algorithm (IPPO) **(author?)** (37) for independent yet collaborative agent actions. Both the RL algorithms converged to policies with similar performance.

## Proposed Solution

Based on the above formulation for the MDPs for Load Shifting $MDP_{LS}$, Energy Reduction $MDP_E$, and Battery Operation $MDP_{BAT}$ , we outline the DC Carbon Footprint Reduction (DC-CFR) multi-agent Reinforcement Learning algorithm (MARL) approach. The main goal is to efficiently reduce the total DC carbon footprint in real time by solving the MDPs simultaneously.

We undertake a systematic implementation of the multi-agent algorithm that accounts for the interdependency between individual agent actions and the next states.

**Input**: The operational/simulation diagram of the approach is described in detail in Figure 2. We simultaneously initialize the three DRL **control agents** $A_{LS}$, $A_E$ and $A_{BAT}$ for the optimal load shifting ($MDP_{LS}$), optimizing energy ($MDP_E$) and the optimal battery operation ($MDP_{BAT}$ ) respectively. Grid data, which includes grid energy and carbon distribution, weather conditions, and compute load, is configured to be queried from a database at every time step. On the other hand, the variables like DC temperature, IT Load, Unassigned flexible load, DC Energy, and Battery state of charge information are obtained via the exchange of information between the individual MDPs as a part of the **Data Center Simulator** which has the Load Shifting Model (2), Energy Plus model for the data center thermodynamics (6) and Battery Storage Model (2; 35; 15; 16). The information exchange processes occur through the **RL Interface** with Open AI Gym wrappers. We shall provide detailed steps for this process when we describe the rollout stage for the agents.

**Rollout**: This is the real-time component of the approach. The agents $A_{LS}$, $A_E$ and $A_{BAT}$ are allowed to interact with their respective MDPs to collect rollout information $(S_t, A_t, S_t, R_{t+1}, \gamma, \text{done})$ in their respective memory buffers D. The different stages of the information exchange between MDPs are captured in the interdependency Fig. 3.

In the beginning, the agent $A_{LS}$ considers the state variables time and CI. Unassigned flexible load is obtained from the $MDP_{LS}$, DC temperature, IT Load, and DC Energy are obtained from $MDP_E$, and Battery state of charge information is obtained from $MDP_{BAT}$ . $A_{LS}$ uses this information to decide its action on whether to reassign flexible load from this instant or to stay idle. The resulting IT load information is passed to the energy-optimizing agent $A_E$. It uses time as well as DC temperature, IT Load, DC Energy, and HVAC Setpoint obtained from the previous time step of $MDP_E$ to decide the setpoint for the next time interval. The Energy Plus model of the DC calculates the resulting changes in energy consumption and DC temperature, and then communicates these back to $MDP_E$. Finally, the agent $A_{BAT}$ considers the time, DC Energy from $MDP_E$, the current battery charge, and the CI information to decide on charging the battery from the grid or supplementing DC energy demand.

From an implementation perspective, the individual agents do not receive the rollout information tuple immediately after taking an action. They wait until it is their turn to take the action again. This incorporates the effect of its action in all other MDPs, making the reward more informed. We use a collaborative reward that considers the effects of actions from all agents. This formulation has been highlighted in Table 1. This enables the agents to look at the effects of their individual actions across dependent MDPs.

**Concurrent Policy Update**: At regular intervals, the buffer data is used to update the agent policies. For training the RL agents in this work, we are primarily using PPO (37). Any other off-policy or on-policy agent may be used.

The overall DC-CFR approach is summarized in Algorithm 1.

## Experiments

We conducted our experiments using EnergyPlus, an open-source building energy simulation software that can simulate the thermal performance of buildings and cooling systems. A two-zone DC HVAC with economizer model was used to simulate a DC consisting of two isolated zones with servers and HVAC cooling. We connected Python with EnergyPlus using the Sinergym framework (9), which wraps the EnergyPlus simulation engine following the OpenAI Gym interface to develop our control algorithms using DRL. This allows us to do step by step simulation of a DC and to dynamically change the cooling setpoint and the running workload of the

| | **$MDP_{LS}$**<br>Flexible Load Shifting | **$MDP_E$**<br>Energy HVAC Optimizer | **$MDP_{BAT}$**<br>Battery Agent |
|---|---|---|---|
| State: $S_t$ | Time, DC temperature, IT Load, Unassigned Flexible Load, DC Energy, Carbon Intensity, Battery Charge | Time, DC temperature, Weather, DC Energy, IT Load, HVAC Setpoint | Time, DC Energy, Battery Charge, Carbon Intensity |
| Action: $A_t$ | Assign Flexible Load, Idle | HVAC Setpoint | Charge, Supply, Idle |
| Reward: $R_{t+1}(S_t, A_t)$ | $0.8 * r_{LS} + 0.1 * r_E + 0.1 * r_{BAT}$ | $0.1 * r_{LS} + 0.8 * r_E + 0.1 * r_{BAT}$ | $0.1 * r_{LS} + 0.1 * r_E + 0.8 * r_{BAT}$ |

Table 1: MDPs for Load Shifting, HVAC Energy Optimization, and Battery Operation. Here $r_{LS} = -(CO_2 \text{ Footprint} + LS_{Penalty})$, $r_E = -(\text{Total Energy Consumption} \times \text{Cost per kWh})$, and $r_{BAT} = -(CO_2 \text{ Footprint})$, where $LS_{Penalty}$ is the scalar value of the unassigned flexible IT workload.

---

Algorithm 1: Data Center Carbon Footprint Reduction DC-CFR Multi-Agent Algorithm

---

**Require:** RL Agents $A_{LS}$, $A_E$ and $A_{BAT}$ ▷ RL Algorithm initialization
**Require:** CI ▷ Carbon Intensity data from the grid
**Require:** EW ▷ Weather data obtained from EnergyPlus
**Require:** Workload Model $MDP_{LS}$ ▷ Model data center workload assignment
**Require:** Data Center Model $MDP_E$ ▷ Model Data Center Thermodynamics in Energy Plus
**Require:** Battery Model $MDP_{BAT}$ ▷ Model Battery operation
**for** i $\in 1, \ldots, L_b$ **do** ▷ $L_b$ is the learning iterations budget
    ***Concurrent Rollout Phase***
    **while** *episode not done* **do**
        State information is shared among the different MDPs
        *Agent* $A_{LS}$ *sends action to* $MDP_{LS}$ *and collects* $(s_t, a_t, s_{t'}, r_t, \gamma, \text{done})$ *in its replay buffer* $D_{LS}$
        *Agent* $A_E$ *sends action to* $MDP_E$ *and collects* $(s_t, a_t, s_{t'}, r_t, \gamma, \text{done})$ *in its replay buffer* $D_E$
        *Agent* $A_{BAT}$ *sends action to* $MDP_{BAT}$ *and collects* $(s_t, a_t, s_{t'}, r_t, \gamma, \text{done})$ *in its replay buffer* $D_{BAT}$
    **end while**
    ***Concurrent Policy Updates***
    *Update Agent Networks by training* $A_E$ *on* $D_E$, $A_{LS}$ *on* $D_{LS}$ *and* $A_{BAT}$ *on* $D_{BAT}$
**end for**

---

DC. The load shifting and the battery models are based on work done in (2). These models are similarly wrapped via Open AI Gym interface.

For $A_{LS}$, as shown in (2), we set the flexible workload to constitute 10% of the DC's total daily workload. Moreover, the server capacity is at each time step is limited, preventing the assignment of all workloads in a single time slot. For $A_{BAT}$ , we assume an installed battery capacity of 50% of DC max hourly energy consumption, as can be found in the uninterrupted power supply (UPS).

Our solution is designed with a reward signal that motivates the agents to reduce both energy consumption, carbon footprint and cost of energy. We have set the action interval at 15-minute time-step, which enables precise control of the system and to quickly respond to changes in the DC environment. We used IT load data of a large-scale real-world DC from the Alibaba (5) open source data set to improve the representativeness of our simulation.

We used New York weather and CI data to train our agents. To improve the generalizability of our solution, we employ an Ornstein-Uhlenbeck (OU) (7) process to introduce noise into the weather data.

We tested the generality of our trained agents by evaluating their performance under diverse climatic and CI conditions. This was done using weather and CI data from three different locations: Arizona (AZ), New York (NY), and Washington (WA). These weather and CI files correspond to locations with distinct weather patterns, ranging from hot and arid to cold and humid. Additionally, we considered Time-of-Use rate plans for energy cost, where the cost vary with the hour.

By validating the model on various locations and weather conditions, we demonstrate the effectiveness of our approach in handling diverse environmental scenarios.

## Experimental Setup

For training our agents, the Rllib (10) implementation of PPO (37) was employed. The hyperparameters used for our experiments are the following: LR = $5 \times 10^{-5}$; Entropy Coefficient = 0.05; Clip Parameter = 0.05; $\gamma = 0.99$; $\lambda$ (GAES Coefficient) = 0.95. The grid search function from Ray Tune (11) was used to find the best learning rate, entropy coefficient and clip values. All agents use a neural network with 3 hidden layers of 128, 64 and 16 units each. The total computing budget for our experiments was approximately 1000 compute hours, utilizing 48 Intel(R) Xeon(R) Gold 6248 CPU @ 2.50GHz cores at an average utilization of less than 50%.

| | **Percentage Reduction of Carbon Footprint** with IPPO compared to ASHRAE<br>Data Center Max Load 1.2MWh<br>Experiment with EnergyPlus for a period of 1 year; Lookahead N = 4 hours | | | | | | |
|---|---|---|---|---|---|---|---|
| | Algorithms | | | | | | |
| | LS | EO | BAT | LS+EO | LS+BAT | EO+BAT | DC-CFR<br>(Our proposal) |
| Arizona | 7.72 ± 0.18 | 8.16 ± 0.05 | 0.25 ± 0.08 | 13.26 ± 0.07 | 7.98 ± 0.1 | 8.46 ± 0.05 | 14.36 ± 0.09 |
| New York | 7.13 ± 0.19 | 8.02 ± 0.06 | 0.41 ± 0.03 | 14.39 ± 0.08 | 7.68 ± 0.20 | 8.21 ± 0.07 | 15.08 ± 0.11 |
| Washington | 4.27 ± 0.20 | 7.54 ± 0.11 | 0.46 ± 0.05 | 13.62 ± 0.08 | 4.53 ± 0.17 | 7.78 ± 0.08 | 13.96 ± 0.06 |

Table 2: Carbon Footprint Reduction Percentages compared to industry standard ASHRAE: Performance of the individual approaches over a period of one year. We are ignoring embodied footprint for server and battery manufacturing.

| | **Percentage Reduction of Energy Consumption** with IPPO compared to ASHRAE<br>Data Center Max Load 1.2MWh<br>Experiment with EnergyPlus for a period of 1 year; Lookahead N = 4 hours | | | | | | |
|---|---|---|---|---|---|---|---|
| | Algorithms | | | | | | |
| | LS | EO | BAT | LS+EO | LS+BAT | EO+BAT | DC-CFR<br>(Our proposal) |
| Arizona | 7.11 ± 0.17 | 8.32 ± 0.04 | 0.00 ± 0.00 | 14.28 ± 0.07 | 7.15 ± 0.09 | 8.41 ± 0.05 | 14.54 ± 0.33 |
| New York | 7.05 ± 0.18 | 8.07 ± 0.06 | 0.00 ± 0.00 | 14.35 ± 0.08 | 7.12 ± 0.20 | 8.28 ± 0.08 | 14.62 ± 0.07 |
| Washington | 4.38 ± 0.21 | 7.42 ± 0.11 | 0.00 ± 0.00 | 13.78 ± 0.06 | 4.46 ± 0.18 | 7.31 ± 0.04 | 13.85 ± 0.07 |

Table 3: Energy Reduction Percentages compared to industry standard ASHRAE evaluated over a period of one year.

| | **Percentage Reduction of Energy Cost** with IPPO compared to ASHRAE<br>Data Center Max Load 1.2MWh<br>Experiment with EnergyPlus for a period of 1 year; Lookahead N = 4 hours | | | | | | |
|---|---|---|---|---|---|---|---|
| | Algorithms | | | | | | |
| | LS | EO | BAT | LS+EO | LS+BAT | EO+BAT | DC-CFR<br>(Our proposal) |
| Arizona | 7.38 ± 0.20 | 8.41 ± 0.04 | 0.23 ± 0.07 | 13.81 ± 0.11 | 7.59 ± 0.10 | 8.43 ± 0.05 | 14.07 ± 0.17 |
| New York | 6.74 ± 0.18 | 8.17 ± 0.06 | 0.31 ± 0.04 | 13.61 ± 0.09 | 7.66 ± 0.22 | 8.39 ± 0.07 | 14.16 ± 0.11 |
| Washington | 3.57 ± 0.17 | 7.52 ± 0.11 | 0.30 ± 0.02 | 12.81 ± 0.05 | 3.81 ± 0.14 | 7.32 ± 0.05 | 12.85 ± 0.05 |

Table 4: Energy Cost Reduction Percentages compared to industry standard ASHRAE evaluated over a period of one year.

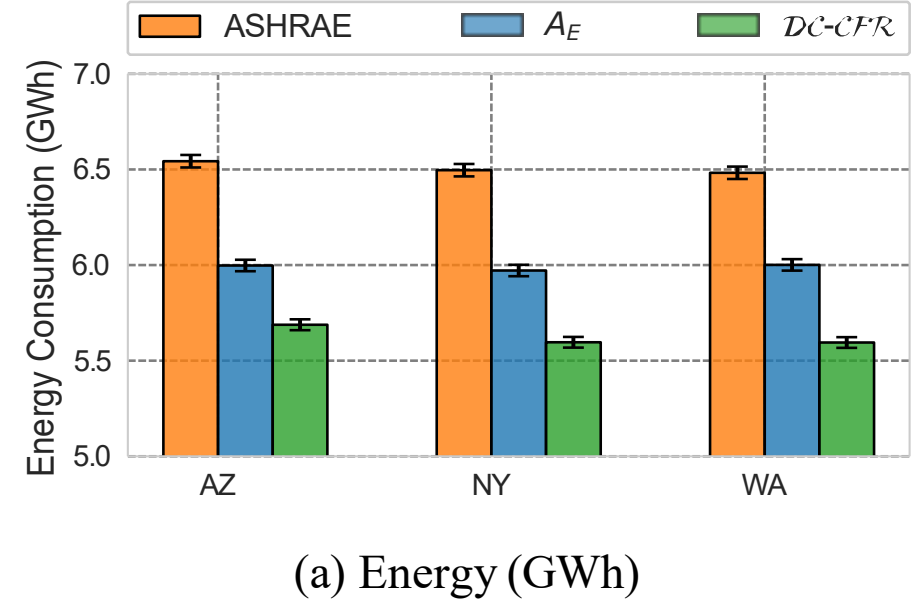

(a) Energy (GWh)

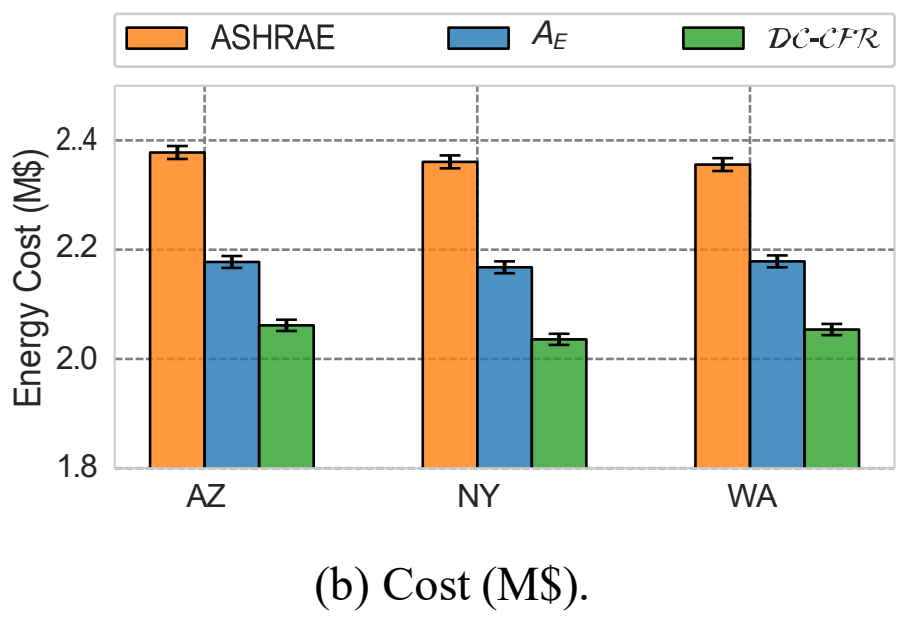

(b) Cost (M$).

Figure 4: Summary of results for data center (DC) control approaches in a 1.2 MWh DC in different locations. ASHRAE is an industry-standard controller for HVAC in DCs.

## Results

In this section, we present the results of our approach evaluated against ASHRAE, the state-of-the-art controller used in DC cooling. To test the robustness of our approach and obtain more diverse results, we ran each experiment 20 times with different random seeds with varying noise in the weather data.

Tables 2, 3, and 4 present the annual reductions in Carbon Footprint, Energy Consumption, and Energy Cost, respectively, that can be achieved by using the DC-CFR framework and various combinations of the agents. These reductions were evaluated in three different locations. Importantly, the comprehensive DC-CFR approach outperforms the individual strategies. This superior performance is attributed to its ability to leverage the interdependencies among different aspects of data center operations and agents, thereby creating a more effective combined policy for energy and cost optimization.

The results obtained show that the proposed approach is able to achieve a high amount of savings in all the three metrics evaluated. On energy consumption, $A_{BAT}$ has no effect since it cannot directly affect the power consumed by the (7).

Figures 1 and 4 provides a comprehensive summary of DC-CFR results, showcasing the optimization across various metrics and locations (Fig. 1 $CO_2$ footprint, Fig. 4 (a) Total energy consumption, Fig. 4 (b) Total energy cost) compared to ASHRAE and our standalone $A_E$ agent. The figure shows how DC-CFR enhances performance on these metrics relative to the industry-standard ASHRAE.

Figure 6 illustrates how DC-CFR opportunistically increased energy expenditure on HVAC cooling, as shown by "spending". By enhancing cooling, DC-CFR effectively decreases the energy consumption of the IT infrastructure. This lowers the total energy consumption of the DC, as evidenced by the "Savings" in Figure 6.

Figure 5 (a) illustrates the actions of the battery agent ($A_{BAT}$) as it charges during periods of low CI and discharges supplying energy during periods of high CI. Figure 5(b) compares DC-CFR $A_{LS}$'s Carbon Aware flexible IT workload assignment against the default workload. $A_{LS}$ shifts flexible IT load to low grid CI hours.

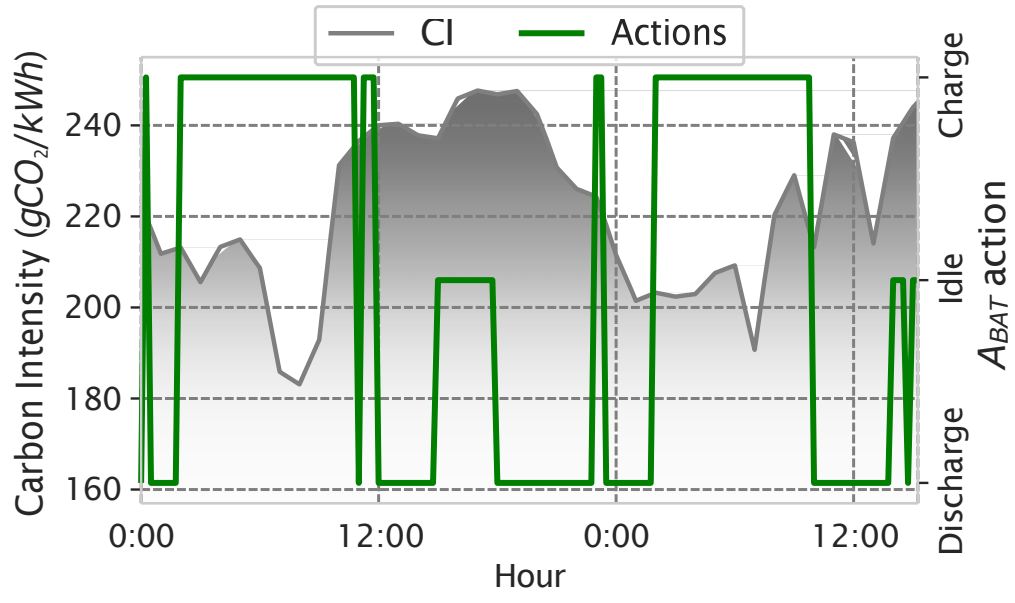

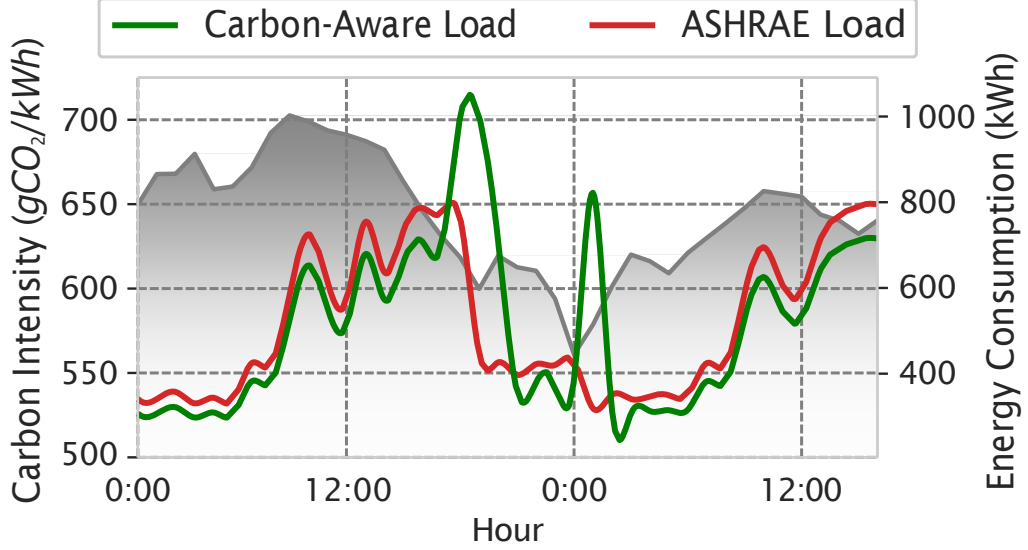

Figure 5: Snapshot of: (top) Actions taken by the $A_{BAT}$ based on CI; (bottom) Carbon Aware Workload (Our proposal) against ASHRAE workload.

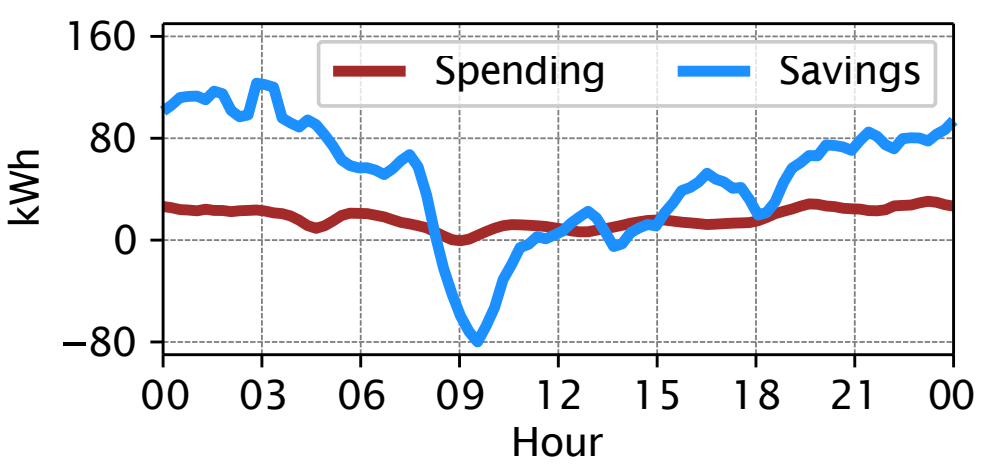

Figure 6: Energy Spending vs Savings over ASHRAE.

## Conclusions

This paper presents a holistic framework called DC Carbon Footprint Reduction (DC-CFR), which optimizes data center energy consumption, load shifting, and battery operation decisions in real-time using Deep Reinforcement Learning (DRL). The framework employs three specialized agents working in concert to substantially reduce both carbon emissions and energy consumption.

The proposed DC-CFR methodology offers significant benefits over traditional static analysis methods. It effectively manages the complex interdependencies among various optimization strategies and uses short-term grid carbon intensity data to guide decision-making. Therefore, unlike static optimizations that rely on long forecast horizons and static seasonal models, our approach can deliver real-time optimization results in dynamic real-world applications.

We have evaluated our approach in multiple data center scenarios across various geographical locations, comparing it with industry-standard solutions such as the ASHRAE rule-based controller. Our method demonstrated significant improvements in carbon footprint reduction, energy efficiency, and cost of energy consumption. DC-CFR is effective for achieving sustainability goals in data center operations.

We plan to open-source the DC-CFR framework with the data center simulation, pluggable control agent abstraction, and OpenAI Gym RL interface, to democratize the carbon reduction efforts by the ecosystem. We plan to further enhance our DC-CFR framework by incorporating other optimization agents across the data center operations like dynamic heterogeneous computation resource allocation for $CO_2$ reduction, and achieving higher QoS with a lower hardware and carbon footprint. We plan to introduce ML CFD surrogates for better heat map estimates (26). This will also enable digital twins for sustainable data centers, and the scalable architecture makes it applicable to supercomputing.

## Follow Up work

The follow-up work was published at **NeurIPS 2024 paper SustainDC** (14), which provides a customizable simulation and benchmark framework in open source to enable ML researchers and Data Center designers. SustainDC provides a highly flexible suite of Python-based Gymnasium environments tailored specifically for multi-agent reinforcement learning (MARL) in data center operations. It allows extensive customization, covering diverse aspects including workloads, server configurations, cooling systems, and aux-iliary battery management, enabling detailed evaluation of various DC design choices. The other short papers on these topics are in the AAAI'24 proceedings (34) and, more notably, in the AAAI'25 proceedings on hierarchical data centers (32).

## References

[1] Sara Abedi, Sang Won Yoon, and Soongeol Kwon. Battery energy storage control using a reinforcement learning approach with cyclic time-dependent markov process. *International Journal of Electrical Power & Energy Systems*, 134:107368, Jan. 2022.

[2] Bilge Acun, Benjamin Lee, Fiodar Kazhamiaka, Kiwan Maeng, Udit Gupta, Manoj Chakkaravarthy, David Brooks, and Carole-Jean Wu. Carbon explorer: A holistic framework for designing carbon aware datacenters. In *Proceedings of the 28th ACM International Conference on Architectural Support for Programming Languages and Operating Systems, Volume 2*. ACM, Jan. 2023.

[3] Marco Biemann, Fabian Scheller, Xiufeng Liu, and Lizhen Huang. Experimental evaluation of model-free reinforcement learning algorithms for continuous HVAC control. *Applied Energy*, 298:117164, Sept. 2021.

[4] Jun Cao, Dan Harrold, Zhong Fan, Thomas Morstyn, David Healey, and Kang Li. Deep reinforcement learning-based energy storage arbitrage with accurate lithium-ion battery degradation model. *IEEE Transactions on Smart Grid*, 11(5):4513–4521, Sept. 2020.

[5] Yue Cheng, Zheng Chai, and Ali Anwar. Characterizing co-located datacenter workloads: An alibaba case study. In *Proceedings of the 9th Asia-Pacific Workshop on Systems*, APSys '18, New York, NY, USA, 2018. Association for Computing Machinery.

[6] Drury B Crawley, Linda K Lawrie, Curtis O Pedersen, and Frederick C Winkelmann. Energy plus: energy simulation program. *ASHRAE journal*, 42(4):49–56, 2000.

[7] F. Espen, Benth, and J. Šaltyte-Benth. Stochastic modelling of temperature variations with a view towards weather derivatives, 2005.

[8] Bin Huang and Jianhui Wang. Deep-reinforcement-learning-based capacity scheduling for PV-battery storage system. *IEEE Transactions on Smart Grid*, 12(3):2272–2283, May 2021.

[9] Javier Jime´nez-Raboso, Alejandro Campoy-Nieves, Antonio Manjavacas-Lucas, Juan Go´mez-Romero, and Miguel Molina-Solana. Sinergym: A building simulation and control framework for training reinforcement learning agents. In *Proceedings of the 8th ACM International Conference on Systems for Energy-Efficient Buildings, Cities, and Transportation*, page 319–323, New York, NY, USA, 2021. Association for Computing Machinery.

[10] Eric Liang, Richard Liaw, Robert Nishihara, Philipp Moritz, Roy Fox, Ken Goldberg, Joseph Gonzalez, Michael Jordan, and Ion Stoica. RLlib: Abstractions for distributed reinforcement learning. In Jennifer Dy and Andreas Krause, editors, *Proceedings of the 35th International Conference on Machine Learning*, volume 80 of *Proceedings of Machine Learning Research*, pages 3053–3062. PMLR, 10–15 Jul 2018.

[11] Richard Liaw, Eric Liang, Robert Nishihara, Philipp Moritz, Joseph E. Gonzalez, and Ion Stoica. Tune: A research platform for distributed model selection and training. *CoRR*, abs/1807.05118, 2018.

[12] Ryan Lowe, Yi I Wu, Aviv Tamar, Jean Harb, OpenAI Pieter Abbeel, and Igor Mordatch. Multi-agent actor-critic for mixed cooperative-competitive environments. *Advances in neural information processing systems*, 30, 2017.

[13] Muhammad Haiqal Bin Mahbod, Chin Boon Chng, Poh Seng Lee, and Chee Kong Chui. Energy saving evaluation of an energy efficient data center using a model-free reinforcement learning approach. *Applied Energy*, 322:119392, Sept. 2022.

[14] Avisek Naug, Antonio Guillen, Ricardo Luna, Vineet Gundecha, Cullen Bash, Sahand Ghorbanpour, Sajad Mousavi, Ashwin Ramesh Babu, Dejan Markovikj, Lekhapriya D Kashyap, Desik Rengarajan, and Soumyendu Sarkar. Sustaindc: Benchmarking for sustainable data center control. In *Advances in Neural Information Processing Systems*, volume 37, pages 100630–100669, 2024.

[15] Avisek Naug, Antonio Guillen, Ricardo Luna Gutie´rrez, Vineet Gundecha, Sahand Ghorbanpour, Lekhapriya Dheeraj Kashyap, Dejan Markovikj, Lorenz Krause, Sajad Mousavi, Ashwin Ramesh Babu, and Soumyendu Sarkar. Pydcm: Custom data center models with reinforcement learning for sustainability. In *Proceedings of the 10th ACM International Conference on Systems for Energy-Efficient Buildings, Cities, and Transportation*, BuildSys '23, page 232–235, New York, NY, USA, 2023. Association for Computing Machinery.

[16] Avisek Naug, Antonio Guillen, Ricardo Luna Gutierrez, Vineet Gundecha, Sahand Ghorbanpour, Sajad Mousavi, Ashwin Ramesh Babu, and Soumyendu Sarkar. A configurable pythonic data center model for sustainable cooling and ml integration. In *NeurIPS 2023 Workshop on Tackling Climate Change with Machine Learning*, 2023.

[17] Ana Radovanovic´, Ross Koningstein, Ian Schneider, Bokan Chen, Alexandre Duarte, Binz Roy, Diyue Xiao, Maya Haridasan, Patrick Hung, Nick Care, Saurav Talukdar, Eric Mullen, Kendal Smith, MariEllen Cottman, and Walfredo Cirne. Carbon-aware computing for datacenters. *IEEE Transactions on Power Systems*, 38(2):1270–1280, Mar. 2023.

[18] Yongyi Ran, Han Hu, Yonggang Wen, and Xin Zhou. Optimizing energy efficiency for data center via parameterized deep reinforcement learning. *IEEE Transactions on Services Computing*, pages 1–14, 2022.

[19] Yongyi Ran, Han Hu, Xin Zhou, and Yonggang Wen. DeepEE: Joint optimization of job scheduling and cooling control for data center energy efficiency using deep reinforcement learning. In *2019 IEEE 39th International Conference on Distributed Computing Sys-*

*tems (ICDCS)*. IEEE, July 2019.

[20] Yongyi Ran, Xin Zhou, Han Hu, and Yonggang Wen. Optimizing data centre energy efficiency via event driven deep reinforcement learning. In *2022 IEEE World Congress on Services (SERVICES)*. IEEE, July 2022.

[21] Soumyendu Sarkar, Ashwin Ramesh Babu, Vineet Gundecha, Antonio Guillen, Sajad Mousavi, Ricardo Luna, Sahand Ghorbanpour, and Avisek Naug. Rl-cam: Visual explanations for convolutional networks using reinforcement learning. In *Proceedings of the IEEE/CVF Conference on Computer Vision and Pattern Recognition (CVPR) Workshops*, pages 3860–3868, June 2023.

[22] Soumyendu Sarkar, Ashwin Ramesh Babu, Vineet Gundecha, Antonio Guillen, Sajad Mousavi, Ricardo Luna, Sahand Ghorbanpour, and Avisek Naug. Robustness with query-efficient adversarial attack using reinforcement learning. In *Proceedings of the IEEE/CVF Conference on Computer Vision and Pattern Recognition*, pages 2329–2336, 2023.

[23] Soumyendu Sarkar, Ashwin Ramesh Babu, Sajad Mousavi, Zachariah Carmichael, Vineet Gundecha, Sahand Ghorbanpour, Ricardo Luna Gutierrez, Antonio Guillen, and Avisek Naug. Benchmark generation framework with customizable distortions for image classifier robustness. In *Proceedings of the IEEE/CVF Winter Conference on Applications of Computer Vision*, pages 4418–4427, 2024.

[24] Soumyendu Sarkar, Ashwin Ramesh Babu, Sajad Mousavi, Sahand Ghorbanpour, Vineet Gundecha, Ricardo Luna Gutierrez, Antonio Guillen, and Avisek Naug. Reinforcement learning based black-box adversarial attack for robustness improvement. In *2023 IEEE 19th International Conference on Automation Science and Engineering (CASE)*, pages 1–8. IEEE, 2023.

[25] Soumyendu Sarkar, Ashwin Ramesh Babu, Sajad Mousavi, Vineet Gundecha, Sahand Ghorbanpour, Alexander Shmakov, Ricardo Luna Gutierrez, Antonio Guillen, and Avisek Naug. Robustness with black-box adversarial attack using reinforcement learning. In *AAAI 2023: Proceedings of the Workshop on Artificial Intelligence Safety 2023 (SafeAI 2023)*, volume 3381. https://ceur-ws.org/Vol-3381/8.pdf, 2023.

[26] Soumyendu Sarkar, Antonio Guillen, Zachariah Carmichael, Vineet Gundecha, Avisek Naug, Ashwin Ramesh Babu, and Ricardo Luna Gutierrez. Enhancing data center sustainability with a 3d cnn-based cfd surrogate model. In *NeurIPS 2023 Workshop on Tackling Climate Change with Machine Learning*, 2023.

[27] Soumyendu Sarkar, Vineet Gundecha, Sahand Ghorbanpour, Alexander Shmakov, Ashwin Ramesh Babu, Alexandre Pichard, and Mathieu Cocho. Skip training for multi-agent reinforcement learning controller for industrial wave energy converters. In *2022 IEEE 18th International Conference on Automation Science and Engineering (CASE)*, pages 212–219. IEEE, 2022.

[28] Soumyendu Sarkar, Vineet Gundecha, Sahand Ghorbanpour, Alexander Shmakov, Ashwin Ramesh Babu, Avisek Naug, Alexandre Pichard, and Mathieu Cocho. Function approximation for reinforcement learning controller for energy from spread waves. In Edith Elkind, editor, *Proceedings of the Thirty-Second International Joint Conference on Artificial Intelligence, IJCAI-23*, pages 6201–6209. International Joint Conferences on Artificial Intelligence Organization, 8 2023. AI for Good.

[29] Soumyendu Sarkar, Vineet Gundecha, Alexander Shmakov, Sahand Ghorbanpour, Ashwin Ramesh Babu, Paolo Faraboschi, Mathieu Cocho, Alexandre Pichard, and Jonathan Fievez. Multi-objective reinforcement learning controller for multi-generator industrial wave energy converter. In *NeurIPs Tackling Climate Change with Machine Learning Workshop*, 2021.

[30] Soumyendu Sarkar, Vineet Gundecha, Alexander Shmakov, Sahand Ghorbanpour, Ashwin Ramesh Babu, Paolo Faraboschi, Mathieu Cocho, Alexandre Pichard, and Jonathan Fievez. Multi-agent reinforcement learning controller to maximize energy efficiency for multi-generator industrial wave energy converter. In *Proceedings of the AAAI Conference on Artificial Intelligence*, volume 36, pages 12135–12144, 2022.

[31] Soumyendu Sarkar, Sajad Mousavi, Ashwin Ramesh Babu, Vineet Gundecha, Sahand Ghorbanpour, and Alexander K Shmakov. Measuring robustness with black-box adversarial attack using reinforcement learning. In *NeurIPS ML Safety Workshop*, 2022.

[32] Soumyendu Sarkar, Avisek Naug, Antonio Guillen, Vineet Gundecha, Ricardo Luna Gutie´rrez, Sahand Ghorbanpour, Sajad Mousavi, Ashwin Ramesh Babu, Desik Rengarajan, and Cullen Bash. Hierarchical multi-agent framework for carbon-efficient liquid-cooled data center clusters. *Proceedings of the AAAI Conference on Artificial Intelligence*, 39(28):29694–29696, Apr. 2025.

[33] Soumyendu Sarkar, Avisek Naug, Antonio Guillen, Ricardo Luna Gutierrez, Sahand Ghorbanpour, Sajad Mousavi, Ashwin Ramesh Babu, and Vineet Gundecha. Concurrent carbon footprint reduction (c2fr) reinforcement learning approach for sustainable data center digital twin. In *2023 IEEE 19th International Conference on Automation Science and Engineering (CASE)*, pages 1–8, 2023.

[34] Soumyendu Sarkar, Avisek Naug, Antonio Guillen, Ricardo Luna, Vineet Gundecha, Ashwin Ramesh Babu, and Sajad Mousavi. Sustainability of data center digital twins with reinforcement learning. *Proceedings of the AAAI Conference on Artificial Intelligence*, 38(21):23832–23834, Mar. 2024.

[35] Soumyendu Sarkar, Avisek Naug, Antonio Guillen, Ricardo Luna Gutierrez, Vineet Gundecha, Sahand Ghorbanpour, Sajad Mousavi, and Ashwin Ramesh Babu. Sustainable data center modeling: A multi-agent reinforcement learning benchmark. In *NeurIPS 2023 Workshop on Tackling Climate Change with Machine Learning*, 2023.

[36] Soumyendu Sarkar, Avisek Naug, Ricardo Luna Gutierrez, Antonio Guillen, Vineet Gundecha, Ashwin Ramesh Babu, and Cullen Bash. Real-time carbon footprint minimization in sustainable data centers with reinforcement learning. In *NeurIPS 2023 Workshop on Tackling Climate Change with Machine Learning*, 2023.

[37] John Schulman, Filip Wolski, Prafulla Dhariwal, Alec Radford, and Oleg Klimov. Proximal policy optimization algorithms, 2017.

[38] Alexander Shmakov, Avisek Naug, Vineet Gundecha, Sahand Ghorbanpour, Ricardo Luna Gutierrez, Ashwin Ramesh Babu, Antonio Guillen, and Soumyendu Sarkar. Rtdk-bo: High dimensional bayesian optimization with reinforced transformer deep kernels. In *2023 IEEE 19th International Conference on Automation Science and Engineering (CASE)*, pages 1–8. IEEE, 2023.

[39] Ruihang Wang, Xinyi Zhang, Xin Zhou, Yonggang Wen, and Rui Tan. Toward physics-guided safe deep reinforcement learning for green data center cooling control. In *2022 ACM/IEEE 13th International Conference on Cyber-Physical Systems (ICCPS)*. IEEE, May 2022.

[40] Deliang Yi, Xin Zhou, Yonggang Wen, and Rui Tan. Toward efficient compute-intensive job allocation for green data centers: A deep reinforcement learning approach. In *2019 IEEE 39th International Conference on Distributed Computing Systems (ICDCS)*. IEEE, July 2019.

[41] Kuo Zhang, Peijian Wang, Ning Gu, and Thu D. Nguyen. Greendrl: Managing green datacenters using deep reinforcement learning. In *Proceedings of the 13th Symposium on Cloud Computing*. ACM, Nov. 2022.

[42] Qingang Zhang, Chin-Boon Chng, Kaiqi Chen, Poh-Seng Lee, and Chee-Kong Chui. DRL-s: Toward safe real-world learning of dynamic thermal management in data center. *Expert Systems with Applications*, 214:119146, Mar. 2023.

[43] Qingang Zhang, Wei Zeng, Qinjie Lin, Chin-Boon Chng, Chee-Kong Chui, and Poh-Seng Lee. Deep reinforcement learning towards real-world dynamic thermal management of data centers. *Applied Energy*, 333:120561, Mar. 2023.

[44] Kunshu Zhou, Kaile Zhou, and Shanlin Yang. Reinforcement learning-based scheduling strategy for energy storage in microgrid. *Journal of Energy Storage*, 51:104379, July 2022.